\documentclass{article}

\usepackage[preprint]{neurips_2025}

\usepackage[utf8]{inputenc}
\usepackage[T1]{fontenc}
\usepackage{hyperref}
\usepackage{url}
\usepackage{booktabs}
\usepackage{amsfonts}
\usepackage{nicefrac}
\usepackage{microtype}
\usepackage{xcolor}
\hypersetup{colorlinks=true,linkcolor=blue!50!black,citecolor=blue!50!black,urlcolor=blue!50!black}
\makeatletter\renewcommand{\@noticestring}{Preprint.}\makeatother

\usepackage{multirow}
\usepackage{graphicx}
\usepackage{caption}
\usepackage{float}
\usepackage{amsmath}
\usepackage{amssymb}
\usepackage{wrapfig}
\usepackage{adjustbox}

\usepackage{arydshln}

\title{DIFFNAS: Predictor-Free Neural Architecture Search via Diffusion Model and Zero-Cost Evaluation}

\author{%
  Yize Sun\textsuperscript{1,2,$*$} \enspace
  Hanzhang Ma\textsuperscript{3,$*$} \enspace
  Zixin Wu\textsuperscript{4} \enspace
  Xuechun Li\textsuperscript{1} \enspace
  Volker Tresp\textsuperscript{1,2} \enspace
  Yunpu Ma\textsuperscript{1,2} \\[0.6em]
  \textsuperscript{1}Ludwig-Maximilians-Universit\"at M\"unchen (LMU Munich), Germany \\
  \textsuperscript{2}Munich Center for Machine Learning (MCML), Germany \\
  \textsuperscript{3}Paderborn University, Germany \\
  \textsuperscript{4}Faculty of Computer Science, University of Duisburg-Essen, Germany \\[0.3em]
  {\small $^{*}$Equal contribution. Correspondence to: Yunpu Ma \texttt{<cognitive.yunpu@gmail.com>}.}
}

\begin{document}

\maketitle

\begin{abstract}
Neural Architecture Search (NAS) has advanced the automated design of deep networks, but remains computationally expensive due to reliance on model training or performance predictors. We introduce DIFFNAS, a predictor-free NAS framework that leverages a discrete graph diffusion model guided by zero-cost performance proxies. Unlike prior diffusion-based NAS methods that require learned performance predictors, DIFFNAS conditions the generative process on training-free metrics such as MeCo and SWAP, enabling efficient architecture search with minimal computational overhead. Experiments on NAS-Bench-201 show that DIFFNAS finds the optimal architectures for CIFAR-10 and CIFAR-100 (94.37\% and 73.51\%) without a pretrained predictor, while reducing search cost by up to 17.5× compared to DiffusionNAG. On the MobileNetV3 search space, its accuracy is lower than that of DiffusionNAG. These results highlight the promise of diffusion models combined with zero-cost proxies for scalable and effective architecture search.
\end{abstract}

\section{Introduction}
Designing efficient neural network architectures plays an important role in modern deep learning, but it remains a compute-intensive task~\citep{peng2022pre}. Neural Architecture Search (NAS) can automate the process but also typically has high computational cost, with numerous methods needing either thousands of candidate architectures trained or tested~\citep{elsken2019neural}. Even predictor-based methods attempt to save this cost by learning performance predictors, with those methods also requiring significant amounts of training and adding new sources of bias and error. The fundamental problem is in the search for performing architectures in a resource-efficient way, independent of costly evaluations or surrogate model training.

Existing methods for NAS are broadly classified into predictor-based~\citep{wen2020neural, wei2022npenas}, training-free zero-shot methods~\citep{mellor2021neural, jiang2023meco, peng2024swap}, and diffusion process-based generative models~\citep{an2023diffusionnag}. Predictor-based methods learn a proxy model that approximates the performance of potential architectures, but this lowers search cost in exchange for an increased burden of training and labeled data. Training-free methods employ handcrafted metrics, such as gradient magnitudes or feature correlation statistics, to estimate architecture quality without requiring training, but their performance is task- and search space-dependent. More recently, diffusion-based methods have demonstrated potential in representing the distribution of top-scoring architectures~\citep{an2023diffusionnag, asthana2024multi}. Nonetheless, those methods usually need to train complicated dataset-informed predictors for conducting the generative process, rendering scalability and generalizability challenging.

In this work, we propose DIFFNAS, a predictor-free NAS framework that combines discrete graph diffusion with zero-cost performance proxies to efficiently generate high-quality architectures. Instead of relying on learned performance predictors, DIFFNAS conditions the generative process directly on training-free metrics, primarily \textbf{MeCo}~\citep{jiang2023meco}, with \textbf{SWAP}~\citep{peng2024swap} as an alternative, which evaluate candidate architectures using a single forward pass through untrained networks. This enables a scalable and lightweight search strategy that maintains high accuracy without expensive model evaluations or predictor training.

Our main contributions are summarized as follows:
\begin{itemize}
    \item We propose \textbf{DIFFNAS}, a predictor-free NAS framework that integrates discrete graph diffusion with zero-cost performance proxies for efficient and scalable architecture generation.
    \item We introduce a \textbf{conditioning mechanism} that incorporates training-free metrics such as MeCo and SWAP into both the denoising and sampling processes.
    \item We enhance the diffusion model with a \textbf{reward-guided fine-tuning stage} using DDPO, aligning generation with task-specific objectives without training any performance predictor.
    \item We demonstrate that DIFFNAS finds the \textbf{optimal architectures} for CIFAR-10 and CIFAR-100 on NAS-Bench-201 while reducing search cost by up to \textbf{17.5$\times$} compared to the baseline (DiffusionNAG), and that its generated architectures are consistently strong, with the highest minimum accuracy among generative methods.
\end{itemize}
\begin{figure}[t!]
  \centering
  \includegraphics[width=\columnwidth]{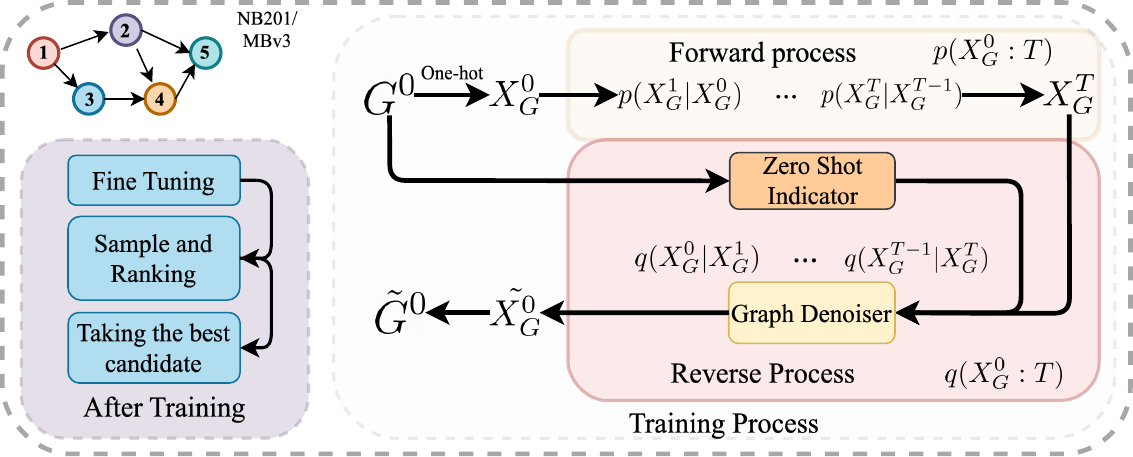}
  \caption{Illustration of diffusion framework and search pipeline for DIFFNAS.}
  \label{fig:diffnas framework}
\end{figure}
\section{Related Work}
\paragraph{Neural Architecture Search} It is an expensive and unintuitive process for human designers to manually construct neural architectures with strong performance for various downstream tasks. NAS aims to automatically find architectures with strong performance after training~\citep{elsken2019neural}. Popular NAS methods can be roughly categorized into Bayesian method-based~\citep{kandasamy2018neural, white2021bananas}, reinforcement learning-based~\citep{zoph2016neural, jaafra2019reinforcement}, gradient-based~\citep{luo2018neural, liu2018darts}, evolutionary algorithm-based~\citep{real2019regularized, lu2020nsganetv2}, predictor-based~\citep{wen2020neural, wei2022npenas}, transferable~\citep{lee2021rapid, an2023diffusionnag}, and training-free NAS~\citep{mellor2021neural, wu2024training}. The first four types of methods can be computationally expensive, since candidate architectures need to be trained and validated to evaluate their performance. Therefore, the predictor-based NAS builds a performance estimator that predicts the performance of candidate architectures to reduce computational cost. However, extra training is required to train predictors to predict the model’s performance accurately based on various downstream tasks. To tackle this challenge, the transferable and training-free NAS investigates designing transferable estimators and training-free indicators. Recently, DiffusionNAG~\citep{an2023diffusionnag} proposed an approach that generates architectures satisfying the objective with a guidance scheme of a meta-learned dataset-aware predictor. However, it still needs extra efforts to train a meta-learned dataset-aware predictor as the model-guidance scheme, while our novel NAS framework eliminates the need for extra predictors by leveraging a diffusion model guided by a zero-cost proxy.

\paragraph{Zero-Shot Indicators} The key component of training-free NAS is defining an indicator
or measurement to estimate the performance of candidate architectures instead of training and validating them on a specific dataset. For example, NASWOT~\citep{mellor2021neural} evaluates and ranks network architectures’ expressivity at zero training cost by applying the Walsh–Hadamard transform to their untrained activations on random inputs and measuring the resulting correlations, ZiCo~\citep{li2023zico} performs zero-shot NAS by measuring the inverse coefficient of variation of gradient norms from untrained networks on random inputs and using more stable gradients as a proxy to rank and select superior architectures, MeCo~\citep{jiang2023meco} uses the minimum eigenvalue of the Pearson correlation matrix of feature maps computed from a single random input’s forward pass through the untrained network as a zero‐cost proxy to evaluate and rank architectures, and SWAP-NAS~\citep{peng2024swap} ranks architectures by using the consistency of untrained networks’ sample-wise activation patterns on input batches as a zero-cost performance proxy. We design a framework incorporating zero-shot indicators as guidance to realize efficient, generalized, and high-performance NAS.

\paragraph{Graph Diffusion Model} Graph diffusion models apply a noise-to-denoise process on graph structures and have driven advances in graph generation (e.g., GraphDDPM~\citep{liu2023generative}, GDSS~\citep{jo2022score}) and GNN message passing (e.g., GDC~\citep{gasteiger2019diffusion}, ADC~\citep{zhao2021adaptive}). Recently, some works have begun to explore graph diffusion for NAS (e.g., DiffusionNAG~\citep{an2023diffusionnag}, Multi-conditioned Graph Diffusion for Neural Architecture Search~\citep{asthana2024multi}).

\section{Methodology}
We illustrate the DIFFNAS framework in Figure~\ref{fig:diffnas framework}. The zero-shot evaluator scores each architecture and passes the score as a conditioning signal to the graph denoiser. After training, the model can optionally be fine-tuned using a reward-guided diffusion objective. During the search phase, candidate architectures are sampled from the latent space conditioned on high scores. These candidates are re-evaluated using the zero-shot metric, ranked, and the top-performing architectures are selected. Unlike other conditional diffusion models for graph generation, which need intensive neural network training, we present a method that uses a training-free metric as reward to guide the reverse process instead of training any predictors.

\paragraph{Neural Architecture Search as Graph Generation}
To formally describe the problem, let us assume that a classical neural network is described as a directed acyclic graph (DAG) $G=(V,E)$, where $V$ stands for nodes corresponding to operation candidates and $E$ stands for edges corresponding to connections of operations. We present the operation matrix as $X_V\in R^{N\times F_V}$ and the adjacent matrix as $X_E \in R^{N\times N \times F_E}$, where $N=|V|$, $F_V$ and $F_E$ stand for the number of nodes, operation types and connection types respectively.

\subsection{Diffusion Process}

Unlike approaches that perturb graphs within a stochastic differential equation (SDE) framework~\citep{jo2022score}, we adopt a discrete diffusion model following DiGress~\citep{vignac2022digress}. The diffusion process is defined as a forward noising procedure on a graph \( G^t = (X_V^t, X_E^t) \), where \( X_V^t \in \mathcal{X}^{n \times d} \) represents node features and \( X_E^t \in \mathcal{E}^{n \times n} \) represents the edge attributes at timestep \( t \). Noise is injected through two transition matrices \( Q_V \in \mathbb{R}^{F_V \times F_V} \) and \( Q_E \in \mathbb{R}^{F_E \times F_E} \), which stochastically replace node and edge tokens with others from the respective marginal distributions.

Formally, at each timestep, the graph is updated via:
\begin{align}
    q(G^t \mid G^{t-1}) = \left(X_V^{t-1} Q_V^t,\; X_E^{t-1} Q_E^t\right),
\end{align}
and the marginal distribution over the noisy graph at timestep \( t \) is given by:
\begin{align}
    q(G^t \mid G^0) = \left(X_V^0 \, \bar{Q}_V^t,\; X_E^0 \, \bar{Q}_E^t\right),
\end{align}
where \( \bar{Q}^t = Q^1 Q^2 \dots Q^t \) is the cumulative transition operator.

The transition matrices \( Q^t \) are chosen to progressively interpolate between the identity matrix and the marginal distribution. Specifically, we set:
\begin{align}
    Q^t = \alpha^t I + (1 - \alpha^t) \mathbf{1} m^\top,
\end{align}
where \( \alpha^t \in [0, 1] \) controls the noise level, \( \mathbf{1} \) is an all-ones column vector, and \( m \) is the marginal distribution over node or edge types. The resulting cumulative transition matrix becomes:
\begin{align}
    \bar{Q}^t = \bar{\alpha}^t I + (1 - \bar{\alpha}^t) \mathbf{1} m^\top,\quad \text{where } \bar{\alpha}^t = \prod_{i=1}^t \alpha^i.
\end{align}
We employ a cosine schedule \citep{nichol2021improved} to control the decay of \( \bar{\alpha}^t \), defined as:
\begin{align}
    \bar{\alpha}^t = \cos^2\left(\frac{\pi (t / T + s)}{2(1 + s)}\right),
\end{align}
with \( s \) as a small offset constant, following standard practice in diffusion models.

Each graph is represented with a unified token matrix \( X_G \) that encodes both nodes and edges, and a corresponding transition matrix \( Q_G \) integrating \( Q_V \) and \( Q_E \) for noise injection in a shared embedding space is applied.

\subsection{Reverse Process}

The reverse process aims to recover the original graph structure \( G^0 \) from a fully corrupted version \( G^T \) by learning a parameterized denoising model that approximates the backward transitions \( p_\theta(G^{t-1} \mid G^t) \). Following DiGress~\citep{vignac2022digress}, we model this process as a Markov chain where the model iteratively predicts less noisy versions of the graph by denoising node and edge tokens.

At each timestep \( t \), we estimate the posterior distribution over the previous state \( G^{t-1} = (X_V^{t-1}, X_E^{t-1}) \) given the current noisy graph \( G^t \), using a learnable model \( f_\theta \) that takes \( G^t \) and \( t \) as inputs. This model outputs a categorical distribution over node and edge types for each token position:
\begin{align}
    p_\theta(G^{t-1} \mid G^t) = \left(p_\theta(X_V^{t-1} \mid G^t, t),\; p_\theta(X_E^{t-1} \mid G^t, t)\right).
\end{align}

We implement \( f_\theta \) using a transformer-based architecture that encodes the entire graph structure, including both nodes and edges, and uses timestep embeddings to incorporate temporal information. The reverse step is viewed as a masked token prediction task, where the model learns to denoise masked or corrupted tokens based on the current graph context.

Training is performed by minimizing the variational bound on the negative log-likelihood of the data, which reduces to a standard cross-entropy loss over the predicted node and edge types:
\begin{align}
    \mathcal{L}_{\text{denoise}} = \mathbb{E}_{t, G^0, G^t} \left[ - \log p_\theta(G^0 \mid G^t, t) \right].
\end{align}
\subsection{Denoiser}

Our denoising model \( f_\theta \) operates on a unified token sequence \( X_G^t \), which jointly represents node and edge tokens at timestep \( t \), and aims to predict the clean graph components \( (X_V^{t-1}, X_E^{t-1}) \) at each reverse diffusion step.

Each token in \( X_G^t \) is enriched with three types of embeddings:
(i) a learned timestep embedding \( \text{Embed}(t) \),
(ii) structural embeddings that encode graph topology, and
(iii) a conditioning signal derived from the evaluator \textsc{MeCo}, namely a binary quality label \( c(G^0) \) obtained by thresholding the MeCo score \( r(G^0) \) of the ground-truth graph \( G^0 \).

In the first training stage, the MeCo score \( r(G^0) \) of each training graph is converted into the binary label \( c(G^0) \): the score is multiplied by 1000 and compared with a threshold of 60,000 (see Experimental Setup). Only architectures above the threshold are kept as training data, and the label is used as an additional conditioning vector in the transformer. Specifically, the label is projected through a linear layer and added to the token representations before passing them to the transformer layers. This provides guidance to the model, encouraging it to reconstruct graphs with high MeCo scores.

To effectively integrate conditioning information, we adopt an adaptive normalization mechanism inspired by AdaLN-Zero~\citep{ede2022adaptive} and FiLM~\citep{perez2018film}. Unlike standard normalization layers (e.g., LayerNorm), our adaptive normalization layer dynamically adjusts the model’s activations based on the conditioning signal. Specifically, given a residual stream \( h \in \mathbb{R}^d \) and a conditioning input \( c \) (e.g., timestep or reward), we apply:
\begin{align}
    \text{AdaLN}(h \mid c) = \gamma(c) \cdot \frac{h - \mu(h)}{\sigma(h)} + \beta(c),
\end{align}
where \( \mu(h), \sigma(h) \) are the feature-wise mean and standard deviation, and \( \gamma(c), \beta(c) \in \mathbb{R}^d \) are scale and shift vectors predicted from \( c \). This allows the model to modulate its behavior in a context-aware manner at every transformer block.

Our transformer backbone consists of alternating self-attention and feed-forward layers. The final output is decoded through two MLP heads for node and edge prediction:
\begin{align}
    \hat{X}_V^{t-1} = \text{MLP}_V(\text{Transformer}(X_G^t, c(G^0))), \quad
    \hat{X}_E^{t-1} = \text{MLP}_E(\text{Transformer}(X_G^t, c(G^0))).
\end{align}

The model is trained using a cross-entropy loss between the predicted and clean node and edge tokens from the original graph:
\begin{align}
    \mathcal{L}_{\text{denoise}} = \mathbb{E}_{t, G^0, G^t} \left[
        \text{CE}(\hat{X}_V^{t-1}, X_V^0) + \text{CE}(\hat{X}_E^{t-1}, X_E^0)
    \right].
\end{align}

By adaptive normalization, our denoiser is equipped to reconstruct high-quality graphs that are both structurally valid and aligned with downstream objectives.

\subsection{Zero-Shot Indicators}

To guide generation toward high-quality structures, we use task-specific evaluators such as MeCo~\citep{jiang2023meco} and SWAP~\citep{peng2024swap}, which provide scalar rewards reflecting the quality of each generated architecture. In the first stage, the scores are thresholded to obtain binary conditioning labels. In the second stage, the continuous scores are used as rewards for DDPO fine-tuning.

Unlike learned property predictors or reward models that require training and backpropagation, MeCo and SWAP are training-free: they compute architectural quality based on simple one-shot structural signals using a single forward pass with no gradients. This enables fast, scalable, and modular evaluation.

\vspace{0.5em}
\noindent \textbf{MeCo}~\citep{jiang2023meco} measures the expressivity of a neural architecture by spectral analysis of its internal feature correlations. For each candidate architecture \( G \), we instantiate a corresponding untrained network \( f_G \) and forward a single random image \( x \sim \mathcal{D} \) through it to obtain the hidden representations at each layer. In layer \( i \), let \( f_i(x; \theta) \in \mathbb{R}^{n \times d} \) denote the feature map. The MeCo score is then defined as the sum of the minimum eigenvalues of the Pearson correlation matrices:
\begin{align}
\text{MeCo}(G) = \sum_{i=1}^{D} \lambda_{\min} \left( P(f_i(x; \theta)) \right),
\end{align}
where \( P(\cdot) \) computes the Pearson correlation matrix across feature dimensions, and \( \lambda_{\min} \) extracts the minimum eigenvalue. A higher score indicates more diverse and expressive internal activations. Furthermore, we introduce another zero-shot indicator SWAP in Appendix~\ref{app:swap}.

\subsection{Reward-Guided Fine-Tuning with DDPO}

While the first-stage diffusion model accurately reconstructs graphs and can be lightly steered by zero-cost scores, its training objective is still maximum likelihood.  
To align the generator directly with the downstream objective, we adopt Denoising Diffusion Policy Optimization (DDPO)~\citep{black2023training} for a second, reward-guided fine-tuning stage.

This second-stage optimization provides an opportunity to align with the task-specific objective by directly optimizing the generation policy for reward, rather than relying solely on supervised reconstruction. We observe that DDPO fine-tuning leads to improvements in search performance across downstream tasks. For more details on DDPO, please refer to Appendix~\ref{app:ddpo}.

\section{Experiment} 
The experiments are designed to evaluate the effectiveness of the proposed method for NAS and to compare its performance with existing approaches.

\subsection{Experimental Setup}
In our experimental evaluation, we detail the search spaces, baseline comparisons, key hyperparameters, and computational resources used to rigorously assess the performance of our proposed framework.
\begin{table*}[t!]
\caption{
Comparison of NAS methods evaluated on the NAS-Bench-201 search space across CIFAR-10, CIFAR-100, Aircraft, and Oxford-IIIT Pets datasets. Methods are grouped by design: Non-weight sharing, Zero-shot, Pre-trained predictor-based NAS, and Diffusion model-based NAS. “A” indicates multiple architecture evaluations during search; “B” indicates use of a pre-trained predictor. “–” denotes unreported results.
}
    \centering
    \begin{adjustbox}{width=\columnwidth,center}
    \begin{tabular}{@{}llcllcc@{}} \toprule 
    \multirow{2}{*}{Method} & \multirow{2}{*}{Search}  & \multicolumn{1}{c}{CIFAR-10} & \phantom{ab} & \multicolumn{1}{c}{CIFAR-100} & \multicolumn{1}{c}{Aircraft} & \multicolumn{1}{c}{Oxford IIIT Pets} \\
    \cmidrule{3-3} \cmidrule{5-7}
    & & Acc. && Acc. & Acc. & Acc. \\
    \midrule
    \multicolumn{7}{c}{\textbf{NAS-Bench-201}} \\
    \midrule
    \multicolumn{7}{c}{\textbf{Non-weight sharing}}\\
    REA~\citep{real2019regularized}       &  - & 93.92 && 71.84 & - & - \\
    RS~\citep{bergstra2012random}        &  - & 93.70 && 71.04 & - & - \\
    REINFORCE~\citep{williams1992simple} &  - & 93.85 && 71.71 & - & - \\
    \midrule
    \multicolumn{7}{c}{\textbf{Zero Shot}}\\
    SynFlow~\citep{tanaka2020pruning}   & A    & 90.40 && 71.10 & - & - \\
    NASWOT~\citep{mellor2021neural}     & A    & 92.96 && 69.98 & - & - \\
    ZiCo~\citep{li2023zico}       & A    & 94.00 && 71.10 & - & - \\
    CVU~\citep{gracheva2021trainless}            & A    & 91.90 && 64.08 & - & - \\
    AZ-NAS~\citep{lee2024az}     & A    & 93.53 && 70.75 & - & - \\
    TE-NAS~\citep{chen2021neural}        & A    & 92.30 && 67.87 & - & - \\
    TG-NAS~\citep{qiao2024tg}  & A    & 93.75 && 70.64 & - & - \\
    Zen-NAS~\citep{lin2021zen}          & A    & 90.60 && 68.10 & - & - \\
    \midrule
    \multicolumn{7}{c}{\textbf{Pre-trained Predictor Based NAS}}\\
    PRE-NAS~\citep{peng2022pre}       & A+B    & 94.04 && 72.02 & - & - \\
    CAP~\citep{ji2024cap}        & A+B    & 94.34 && 73.41 & - & - \\
    RANK-NOSH~\citep{wang2021rank}   & A+B    & 94.26 && 73.51 & - & - \\
    \midrule
    \multicolumn{7}{c}{\textbf{Diffusion Model Based NAS}}\\
    DiffusionNAG~\citep{an2023diffusionnag} & A+B    &94.37 && 73.51 & 58.83 & \textbf{41.80}\\
    \midrule
    DIFFNAS (ours) & A & \textbf{94.37} && \textbf{73.51} & \textbf{58.85} & 41.50\\
\midrule
\end{tabular}
\end{adjustbox}
\begin{flushleft}
    \footnotesize * A and B under Search respectively denote the need for multiple evaluations of architectures and the use of a pre-trained predictor. Results of other methods are taken from the original publications, except for the non-weight-sharing methods~\citep{lee2021rapid}, SynFlow and Zen-NAS (as reported by \citet{li2023zico}), and TE-NAS (as re-evaluated by \citet{lee2024az}).
  \end{flushleft}

\label{tab:nas_comparison_test_copy2}
\end{table*}

\paragraph{Search Space} To validate the efficacy of our framework, we utilize the NAS-Bench-201 (NB201) benchmark search space~\citep{dong2020bench}, which comprises 15,625 unique architectures. Following prior work~\citep{lee2021rapid, an2023diffusionnag}, we report the test accuracy recorded in the benchmark.  We also utilize MobileNetV3 search space, which can generate a total of about $10^{19}$ possible sub-architectures by combining hyperparameters such as convolutional kernel sizes (3, 5, 7), expansion ratios (3, 4, 6), and network depths (2, 3, 4).
\vspace{-0.1in}

\paragraph{Baseline} We compare our method against several generative NAS approaches:
\textbf{DiffusionNAG}~\citep{an2023diffusionnag}: A diffusion-based NAS method guided by a pre-trained predictor that steers the model toward generating high-performance architectures.
\textbf{MetaD2A}~\citep{lee2021rapid} A non-conditional generative model that does not rely on predictors during the architecture generation phase. However, it requires training the generated architectures to obtain their performance metrics, which incurs significant computational overhead. During testing, MetaD2A first generates architectures in an objective and unconditional manner, and subsequently evaluates the top-performing architectures using its predictor.

\vspace{-0.1in}
\paragraph{Hyperparameters} Our approach employs a two-stage training paradigm. In the first stage, a transformer model is trained to learn the characteristics of high-quality architectures. The second stage leverages reward-guided reinforcement learning to steer the framework toward generating architectures with superior performance. The first phase consists of 5000 epochs, while the second phase involves 200 epochs. During the second stage, a reward of -50,000 is applied if the model generates invalid architectures. The framework generates 1000 samples on four datasets: CIFAR-10~\citep{alex2009learning}, CIFAR-100~\citep{alex2009learning}, Aircraft~\citep{maji2013fine}, and Oxford-IIIT Pets~\citep{parkhi2012cats}. We incorporate the zero-shot NAS method, MeCo~\citep{jiang2023meco}, to assign binary labels to neural network structures. To preserve MeCo's accuracy during scaling, we adjust its grading metric by multiplying it by a factor of 1000. A threshold of 60,000 is defined to distinguish between good and poor neural network architectures.

\vspace{-0.1in}
\paragraph{Computational Resources} Experiments were run on the Hawk system (nodes with NVIDIA A100 40 GB GPUs and AMD EPYC 7742 CPUs) and the Hunter system (AMD Instinct MI300A) at HLRS. Architecture search used a single GPU.  Each training (5000 epochs) took 3 GPU-hours on a single GPU with 40 GB VRAM.  DDPO fine-tuning (200 epochs) took 10 GPU-hours per run.  

\subsection{Results on NB201 Search Space}
Table~\ref{tab:nas_comparison_test_copy2} presents the performance of various NAS algorithms on NAS-Bench-201. On \textit{CIFAR-10}, \textbf{DIFFNAS} achieves 94.37\%, matching the best-performing method, DiffusionNAG (94.37\%), while avoiding the need for a pretrained predictor. Among zero-shot baselines, ZiCo reaches 94.00\%, NASWOT achieves 92.96\%, and SynFlow obtains 90.40\%, highlighting a consistent performance gap. On \textit{CIFAR-100}, DIFFNAS attains 73.51\%, outperforming all zero-shot methods (best: ZiCo and SynFlow, 71.10\%) and matching the best pre-trained predictor-based method, RANK-NOSH (73.51\%; CAP reaches 73.41\%). DiffusionNAG reports the same accuracy (73.51\%), but, like RANK-NOSH and CAP, it relies on a learned predictor. In contrast, DIFFNAS combines diffusion-based architecture generation with the MeCo zero-cost proxy, offering competitive performance without the complexity of surrogate modeling.

\begin{table*}[t]
\caption{Comparison of NAS methods on NAS-Bench-201 across CIFAR-10, CIFAR-100, Aircraft, and Oxford-IIIT Pets. Reported metrics include model size (in millions of parameters), search time (GPU hours), monetary cost (in USD), and accuracy (\%). “–” denotes missing or unreported values.}
\vspace{-0.1in}
\small
\begin{center}
\resizebox{\linewidth}{!}{
\begin{tabular}{cl|ccrcc}
\toprule
\multirow{2}{*}{Target Dataset}             & 
\multicolumn{1}{c|}{\multirow{2}{*}{NAS Method}} & 
Params                                      & 
Search Time                                 & 
\multirow{2}{*}{Speed Up}                   &
Search Cost                                 & 
Accuracy                                   \\ 
                                            & 
                                            & 
(M)                                         & 
(GPU h)                                     & 
                                            & 
(\$)                                        & 
(\%)                                        \\
\midrule
\midrule
\multirow{8}{*}{CIFAR-10}                 
& NASWOT             & 0.615  & 0.09  & 22.5×    & 0.22 & 92.96 \\
& TE-NAS             & 1.317  & 0.43  & 107.5×    & 1.07 & 93.78 \\
& TG-NAS             & 0.802  & 0.01  & 2.5×    & 0.02 & 93.75 \\
& DiffusionNAG       & -      & 0.07  & 17.5×    & 0.18 & 94.37 \\
& PC-DARTS           & 1.17   & 2.89  & 722.5× & 7.22 & 93.66 \\
& \textbf{DIFFNAS (Ours)} & – & 0.004 & 1× & \textbf{0.009} & \textbf{94.37} \\
\midrule
\multirow{8}{*}{CIFAR-100}                 
& NASWOT             & 0.615  & 0.09  & 22.5×    & 0.22 & 69.70 \\
& TE-NAS             & 1.317  & 0.43  & 107.5×    & 1.07 & 70.44 \\
& TG-NAS             & 0.802  & 0.01  & 2.5×    & 0.02 & 70.64 \\
& DiffusionNAG       & –      & 0.07  & 17.5×    & 0.18 & 73.51 \\
& PC-DARTS           & 0.26   & 5.54  & 1385× & 13.85 & 66.64 \\
& \textbf{DIFFNAS (Ours)} & – & 0.004 & 1× & \textbf{0.009} & \textbf{73.51} \\
\midrule
\multirow{5}{*}{Aircraft}                 
& RSPS               & 0.22   & 5.19  & 1297.5× & 12.98 & 42.19 \\
& SETN               & 0.44   & 5.16  & 1290× & 12.89 & 44.84 \\
& DiffusionNAG       & –      & 0.07  &  17.5×   & 0.18 & 58.83 \\
& PC-DARTS           & 0.32   & 0.98  & 245×  & 2.45  & 26.33 \\
& \textbf{DIFFNAS (Ours)} & – & 0.004 & 1× & \textbf{0.009} & \textbf{58.85} \\
\midrule
\multirow{5}{*}{Oxford-IIIT Pets}         
& RSPS               & 0.32   & 0.93  & 232.5×  & 2.33 & 22.91 \\
& SETN               & 0.32   & 2.40  & 600× & 5.99  & 25.17 \\
& DiffusionNAG       & –      & 0.07  & 17.5×   & 0.18  & \textbf{41.80} \\
& PC-DARTS           & 0.44   & 0.79  & 197.5×  & 1.97  & 25.31 \\
& \textbf{DIFFNAS (Ours)} & – & 0.004 & 1× & \textbf{0.009} & 41.50 \\
\bottomrule
\end{tabular}
}
\end{center}
\small
\label{tbl:main_nasbench201}
\vspace{-0.1in}
\begin{flushleft}
    \footnotesize
    * \textit{RSPS}~\citep{li2020random}, \textit{SETN}~\citep{dong2019one}, and \textit{PC-DARTS}~\citep{xu2019pc} are NAS methods included for comparison. Values for NASWOT, TE-NAS and TG-NAS are taken from \citet{qiao2024tg}; values for RSPS, SETN and PC-DARTS are taken from \citet{lee2021rapid} and \citet{an2023diffusionnag}, with search times converted from GPU seconds to GPU hours. Accuracies can therefore differ from Table~\ref{tab:nas_comparison_test_copy2}, which uses other sources. Search cost assumes \$2.50 per GPU hour; speed-up is relative to DIFFNAS.
\end{flushleft}
\end{table*}

Table~\ref{tbl:main_nasbench201} reports the performance, efficiency, and cost of various NAS methods on NAS-Bench-201 across four datasets. Metrics include model size, search time (in GPU hours), monetary cost (USD), and test accuracy. DIFFNAS achieves competitive accuracy across the four datasets while reducing search cost by up to 17.5x compared to baseline. On \textit{CIFAR-10} and \textit{CIFAR-100}, it matches DiffusionNAG’s performance (94.37\% and 73.51\%, respectively) while requiring $17.5\times$ less search time (0.004 vs. 0.07 GPU hours), and without the need for a pretrained predictor. Compared to PC-DARTS, which incurs 722.5$\times$ and 1385$\times$ longer search times on the two datasets, DIFFNAS achieves higher accuracy at a fraction of the computational and monetary cost.

On \textit{Aircraft}, DIFFNAS achieves 58.85\%, marginally outperforming DiffusionNAG (58.83\%) while using 17.5$\times$ less compute. It also surpasses RSPS and SETN by more than 15 percentage points, despite their 1000$\times$ higher compute cost. On \textit{Oxford-IIIT-Pet}, DIFFNAS reaches 41.50\%, closely matching DiffusionNAG (41.80\%) with the same $17.5\times$ cost reduction and no predictor overhead.

These results show that DIFFNAS keeps strong accuracy while greatly reducing search cost, which makes it practical for NAS under limited compute budgets.

\begin{table}[t]
    \centering
    \caption{Performance comparison on CIFAR10 and CIFAR100 datasets in NB201 search space. 1000 architectures are generated for computing statistical metrics for each method.}
    \label{tab:NB201-1000}
    \begin{tabular}{@{}llccccc@{}}
        \toprule
        \multirow{2}{*}{\shortstack{Target \\ Dataset}} & 
        \multirow{2}{*}{\shortstack{Stats.}} & 
        \multicolumn{5}{c}{Methods} \\ \cmidrule(lr){3-7}
        & & Oracle & Random & MetaD2A & Diff. & Ours \\ \midrule
        \multirow{3}{*}{CIFAR10}  & Max  & 94.37 & 94.37 & 94.37 & 94.37 & \textbf{94.37} \\
                                  & Mean & 93.50 & 87.12 & 91.52 & 93.13 & \textbf{93.41} \\
                                  & Min  & 93.18 & 10.00 & 10.00 & 86.44 & \textbf{92.08} \\ \midrule
        \multirow{3}{*}{CIFAR100} & Max  & 73.51 & 72.74 & 73.51 & 73.51 & \textbf{73.51} \\
                                  & Mean & 70.62 & 61.59 & 67.14 & \textbf{70.34} & 69.91 \\
                                  & Min  & 69.91 & 1.00  & 1.00  & 58.09 & \textbf{64.39} \\ \bottomrule
    \end{tabular}
\end{table}

In Table \ref{tab:NB201-1000}, we can find the best architectures for both CIFAR10 and CIFAR100 datasets, corresponding to the accuracy of 94.37\% and 73.51\%. On CIFAR-10, our mean and min values are the highest among the generative methods. On CIFAR-100, our min value is the highest, but our mean is slightly lower than that of DiffusionNAG. Notably, the structure corresponding to our min value is significantly higher than the structure corresponding to the other methods, including DiffusionNAG. This indicates that the architectures generated by our method are more consistently good. 

\subsection{Results on MobileNetV3 Search Space}
Table~\ref{tab:Mbv3} in Appendix~\ref{app: mbv3} shows a comparison between DiffusionNAG and our method, DIFFNAS, on CIFAR-10 using the MobileNetV3 search space. While DIFFNAS achieves a lower accuracy (89.50\%) than DiffusionNAG (97.39\%), we attribute this discrepancy primarily to a mismatch between the MeCo zero-shot evaluator and the final training objective in the MobileNetV3 space. MeCo evaluates untrained architectures based on spectral analysis of internal feature correlations, which performs well in broader search spaces like NAS-Bench-201. However, in the MobileNetV3 space, where architectures are highly refined and performance differences stem from subtle interactions between operations and training dynamics, such zero-cost proxies may not effectively capture the fine-grained distinctions that impact downstream accuracy. This misalignment between the proxy signal and true task performance may have limited the effectiveness of MeCo-guided architecture generation in this case.

In addition, the MobileNetV3 search space is known to be sensitive to small architectural changes, making it inherently more difficult for automated methods to discover top-performing designs. Furthermore, due to the absence of publicly available training configurations for DiffusionNAG, we evaluated the architectures discovered by DIFFNAS using self-defined hyperparameters, which may have resulted in suboptimal training and contributed to the performance gap.

While DIFFNAS underperforms on the MobileNetV3 search space, its strong results on other benchmarks suggest that this gap may stem from the unique challenges of MobileNetV3 rather than general weaknesses of the approach.

\subsection{Ablation Studies}

In order to verify the contribution of each module in the framework to the final performance, we designed a series of ablation experiments to remove or replace key components and record their impact on model performance. All experiments are performed in the search space NAS-Bench-201 on the same hardware configuration (NVIDIA A100 GPU) and dataset CIFAR-10.

We conducted comparative experiments by removing or replacing the following components, using the Full Framework as a baseline:
\begin{itemize}
  \item \textbf{w/o MeCo}: Removes the MECO-based reward evaluator; the architecture generation proceeds without binary labeling.
  \item \textbf{MeCo replaced by SWAP}: Substitutes MECO with SWAP, a different predictor-free NAS metric, to assess the impact of MECO-guided labeling.
  \item \textbf{w/o Reward-Guided Phase}: Removes Stage 2 training (DDPO) and trains only with token-level supervision.
  \item \textbf{w/o Diffusion Transformer}: Removes the diffusion-based transformer and retains only baseline components to assess its contribution.
\end{itemize}
\begin{table}[t!]
\centering
\caption{Summary of ablation studies. 1000 architectures are generated for computing statistical metrics for each method.}
\begin{tabular}{@{}llcccccc@{}}
\toprule
\multirow{2}{*}{\shortstack{Target \\ Dataset}} & 
\multirow{2}{*}{\shortstack{Stats.}} & 
\multicolumn{5}{c}{Methods} \\ \cmidrule(lr){3-7}
& & {\shortstack{Oracle \\ Top-1,000}} & {\shortstack{w/o\\MeCo}}& {\shortstack{w/o\\scaling}} & SWAP & {\shortstack{Full \\ Frame}}\\ \midrule
\multirow{3}{*}{CIFAR10}  & Max  & 94.37 & 94.36 & 94.36 & 94.37 & \textbf{94.37} \\
                          & Mean & 93.50 & 87.70 & 92.33 & 86.57 & \textbf{93.41} \\
                          & Min  & 93.18 & 10.00 & 86.64 & 10.00 & \textbf{92.08} \\ \midrule
\end{tabular}
\label{tab:Ablation}
\vspace{-0.1in}
\end{table}
Table~\ref{tab:Ablation} presents the results of ablation studies assessing the contribution of key components in our framework. The Full Framework consistently achieves strong performance, with a maximum accuracy of 94.37\%, a mean accuracy of 93.41\%, and a minimum accuracy of 92.08\%, closely approaching the Oracle Top-1,000 benchmark, which serves as an empirical upper bound.

Removing the MeCo-based binary labeling strategy leads to a notable drop in performance, with mean accuracy decreasing to 87.70\% and minimum accuracy falling to 10.00\%. This highlights MeCo’s importance in stabilizing and guiding architecture generation. Replacing MeCo with SWAP results in even lower performance, with a mean of 86.57\% and a minimum of 10.00\%, indicating that MeCo provides more effective and reliable zero-shot guidance.
Eliminating the reward-guided scaling phase also degrades performance, reducing mean accuracy to 92.33\% and minimum accuracy to 86.64\%. Notably, the maximum accuracy remains high across all ablation settings, suggesting that while strong architectures can still be discovered, the quality of the overall distribution deteriorates when key components are removed.
Finally, when the diffusion transformer is removed entirely, the model fails to generate any valid architectures, highlighting its essential role in structured generation.

These results show that the MeCo evaluator, the reward-guided scaling phase and the diffusion transformer are all needed for consistent, high-quality architecture generation.

\section{Conclusion}
We introduce DIFFNAS, a predictor-free Neural Architecture Search framework that integrates discrete graph diffusion with zero-cost performance proxies. By conditioning the generative process on training-free metrics such as MeCo and SWAP, DIFFNAS enables efficient and scalable architecture generation without the need for learned performance predictors. Experiments on NAS-Bench-201 show that DIFFNAS finds the optimal architectures for CIFAR-10 and CIFAR-100 while substantially reducing search cost compared to existing NAS methods. While performance on MobileNetV3 was lower than prior work, we attribute this to a misalignment between zero-shot proxy signals and true task performance in this more complex search space. Ablation studies show that conditioning, the diffusion-based transformer and reward-guided fine-tuning each contribute to robustness. DIFFNAS thus provides a basis for future work on predictor-free, diffusion-based architecture generation.

\section*{Acknowledgements}
The authors gratefully acknowledge the Gauss Centre for Supercomputing e.V. (\url{www.gauss-centre.eu}) for funding this project by providing computing time on the GCS supercomputers Hawk and Hunter at the High-Performance Computing Center Stuttgart (HLRS).

\bibliographystyle{plainnat}
\bibliography{references}

\clearpage
\appendix
\section{Zero-Shot Indicator SWAP}\label{app:swap}
\vspace{0.5em}
\noindent \textbf{SWAP}~\citet{peng2024swap} quantifies the expressivity of an architecture by measuring the diversity of its activation patterns across different input samples. For a given network \( f_\theta \), we forward a batch of \( S \) inputs \( \{x_1, \dots, x_S\} \) and record the post-activation outputs \( p_s^{(v)} \) at each intermediate unit \( v \) for each sample \( s \). These activations are binarized using an indicator function \( \mathbb{I}[\cdot] \), and the SWAP score is computed as:
\begin{align}
    \Psi_{N, \theta} = \left| \hat{A}_{N, \theta} \right|, \quad \text{where} \quad \hat{A}_{N, \theta} = \left\{ p(v) : p(v) = \mathbb{I}\left[ p_s^{(v)} \right]_{s=1}^{S} \right\}_{v=1}^{V}.
\end{align}
Here, \( p(v) \in \{0,1\}^S \) is the binary activation pattern of unit \( v \) across the batch, and \( \left| \hat{A}_{N, \theta} \right| \) counts the number of unique such patterns. A higher SWAP score indicates that the architecture exhibits more varied responses to different inputs, reflecting greater discriminative capacity.

\vspace{0.5em}
These training-free evaluators are efficient, require no learned components, and reflect important aspects of architectural capacity, making them ideal for use in generative and search-based pipelines.
\section{DDPO}\label{app:ddpo}
Following \citet{black2023training}, each reverse denoising step is viewed as an action in a $T$-step Markov decision process (MDP). Given a trained diffusion model \( p_\theta \), we generate candidate samples \( \{G_i\} \) and score them using the reward function \( r(G) \) based on zero-shot indicators. We use the continuous MeCo score as the reward. When generated samples are invalid, they are assigned a negative penalty reward. Rewards are normalized across each batch of samples before the policy-gradient update. DDPO then fine-tunes the model to maximize the expected reward:
\begin{align}
J(\theta)=\mathbb{E}_{\tau\sim\pi_\theta}\big[r(G)\big],
\end{align}

which we optimise with the score-function (REINFORCE) gradient:
\begin{align}
    \nabla_\theta J(\theta)=
    \mathbb{E}_{\tau\sim\pi_\theta}\!\Bigg[
    \sum_{t=1}^{T}
    \nabla_\theta\log p_\theta(G_{t-1}\!\mid\!G_t,\ c)\,
    \big(r(G_0)\big)
    \Bigg],
\end{align}

\section{MobileNetV3 results}\label{app: mbv3}
\begin{table}[H]
\centering
\caption{Comparison of neural architecture search methods evaluated on the MobileNetV3 search space on CIFAR-10.}
\begin{tabular}{lc}
\toprule
Algorithm & Acc. (CIFAR-10) \\
\midrule
DiffusionNAG & 97.39 \\
DIFFNAS (Ours) & 89.50 \\
\bottomrule
\end{tabular}
\label{tab:Mbv3}
\end{table}
\section{External Asset Licenses}
\label{app:asset-licenses}

To ensure proper credit and compliance, we document below all external assets used in this work, including their versions, licenses, and source URLs.

\begin{table}[H]
  \centering
  \caption{External assets, versions, licenses, and source URLs.}
  \label{tab:asset-license}
  \resizebox{\textwidth}{!}{%

  \begin{tabular}{llll}
    \toprule
    Asset           & Version & License        & Source URL \\
    \midrule
    NAS-Bench-201   & v2.0    & MIT      & \url{https://github.com/D-X-Y/NAS-Bench-201} \\
    MobileNetV3     & 0.1\footnote{According to \citet{cai2020once} version} & Apache-2.0    & \url{https://github.com/mit-han-lab/once-for-all} \\
    MeCo            & 0d830dd\footnote{The latest github commit number}  & MIT            & \url{https://github.com/HamsterMimi/MeCo} \\
    SWAP            & 0853fc8\footnote{The latest github commit number}   & AFL-3.0 license   & \url{https://github.com/pym1024/SWAP} \\
    PyTorch Lightning & 2.5.1 & Apache-2.0     & \url{https://github.com/Lightning-AI/pytorch-lightning} \\
    \bottomrule
  \end{tabular}
  }
\end{table}
\newpage

\end{document}